\documentclass{article}
\usepackage{ijcai22}
\usepackage[left=3cm, right=3cm, top=3cm]{geometry}
\usepackage{times,multicol,enumitem}
\usepackage{soul}
\usepackage{url}
\usepackage[hidelinks]{hyperref}
\usepackage[utf8]{inputenc}
\usepackage[small]{caption}
\usepackage{graphicx}
\usepackage{amsmath}
\usepackage{amsthm}
\usepackage{booktabs}

\usepackage{subfig}
\usepackage{enumitem}
\usepackage{xcolor}
\usepackage{caption}

\usepackage{floatrow}
\usepackage{multirow}
\usepackage{romannum}
\usepackage{algorithm}
\usepackage[noend]{algpseudocode}

\title{Bayesian Neural Hawkes Process for Event Uncertainty Prediction
}

\author{
Manisha Dubey$^1$
\and
Ragja Palakkadavath$^2$\and
P.K. Srijith$^{1}$
\affiliations
$^1$Indian Institute of Technology Hyderabad, India\\
$^2$TCS Research, India\\
\emails
\{cs17resch11003, srijith\}@iith.ac.in,
ragjapk@gmail.com
}

\begin{document}

\maketitle



\begin{abstract}

Event data consisting of time of occurrence of the events arises in several real-world applications. Recent works have introduced neural network based point processes for modeling event-times, and were shown to provide state-of-the-art performance in predicting event-times. However, neural point process models lack a good  uncertainty quantification capability on predictions. A proper uncertainty quantification over event modeling will help in better decision making for many practical applications.  Therefore, we propose a novel point process model, Bayesian Neural Hawkes process (BNHP) which leverages uncertainty modelling capability of Bayesian models and generalization capability of the neural networks to model  event occurrence times. We augment the model with spatio-temporal modeling capability where it can consider uncertainty over predicted time and location of the events. Experiments on simulated and real-world datasets show that BNHP significantly improves prediction performance and uncertainty quantification for modelling events.  
\end{abstract}

\section{Introduction}

Many dynamic systems are characterized by sequences of discrete events over  continuous time. 
Such temporal data are generated in various applications such as earthquakes, taxi-pickups, financial transactions, patient visits to hospitals etc. Despite these diverse domains, a common question across all the platforms is: Given an observed sequence of events, can we predict the time for the occurrence of the next event. 
A practical model should not only be capable of doing prediction but also model uncertainty in its predictions. 
For example, in healthcare predicting the patient arrival and uncertainty around prediction in the form of a time interval can help in queue management in hospitals. Similarly, uncertainty estimates for volcanic eruptions and earthquakes can help in risk management and make proactive decisions. A proper modelling of uncertainty will help users to determine what predictions by the event prediction model one can rely upon. Therefore, we need a model for event time prediction which can capture uncertainty. 
However, the neural models used for predicting temporal events make a point estimation of parameters and lack the required uncertainty modelling capability. This can lead to unreliable and overconfident predictions which can adversely affect decision making in many real-world applications. Therefore, we need to develop models which will also consider \textit{the confidence of the model in making the prediction}. 

Popular approaches for modeling event-data are based on point processes which are defined using a latent intensity function. A Hawkes process (HP)~\cite{hawkes1971spectra} is a point process with self-triggering property i.e. occurrences of previous events trigger occurrences of future events. 
A parametric form of the Hawkes process assumes monotonically decaying influence over past events, which may limit the expressive ability of the model and can deteriorate the performance on complex data sets. Therefore, recent techniques model the intensity function as a neural network~\cite{mei2017neural,du2016recurrent,xiao2017modeling,omi2019fully} so that the influence of past events can be modelled in a non-parametric manner without assuming a particular functional form.

Neural network based models typically tend to yield overconfident predictions and are incapable of properly modelling predictive uncertainty~\cite{gal2016uncertainty}. This can be overcome by combining Bayesian principles in neural networks, leading to Bayesian neural networks (BNN)~\cite{neal2012bayesian}. 
BNN puts a distribution over the weight parameters and computes a posterior over them using Bayes theorem.
Predictions are made using Bayesian model averaging using this posterior distribution which help in modelling uncertainty. However, determination of posterior is fraught with practical difficulties. Monte Carlo (MC) Dropout proposed by~\cite{gal2016dropout,gal2016theoretically} provides a practical and useful approach to develop BNNs. They proved that dropout regularization can be seen as equivalent to performing Bayesian inference over weights of a neural network.

 In this paper, we propose \textit{Bayesian Neural Hawkes Process (BNHP)} where we develop Monte Carlo dropout for a neural Hawkes process consisting of recurrent and feed-forward neural networks. BNHP is not only capable of  modelling uncertainty but also improves the prediction of the event-times.  Our contribution can be summarized as:
\begin{itemize}[leftmargin=10mm]
    \item  We propose a novel model which combines the advantages of the neural Hawkes process and Bayesian neural network for modelling uncertainty over time of occurrence of events. We also extend the proposed approach to spatio-temporal event modeling set-up. 
    \item We develop MC dropout for a neural Hawkes process network which consists of both recurrent and feed-forward neural networks. 
    \item We demonstrate the effectiveness of BNHP for uncertainty modelling and prediction of event times and locations  on several simulated and real world data.  
\end{itemize}

\section{Related Work}
\textbf{Hawkes process:} 
Point processes~\cite{valkeila2008introduction} are useful to model the distribution of points and are defined using an underlying intensity function.  
Hawkes process~\cite{hawkes1971spectra} is a point process \cite{valkeila2008introduction} with self-triggering property. 
i.e occurrence of previous events trigger occurrences of future events. 
Hawkes process has been used in earthquake modelling~\cite{paper20}, crime forecasting~\cite{paper19}, finance~\cite{bacry2015hawkes,embrechts2011multivariate} and epidemic forecasting~\cite{paper18,chiang2021hawkes}. 
However, to avoid parametric models various research works have proposed~\cite{du2016recurrent,mei2017neural,omi2019fully,zuo2020transformer} where intensity function is learnt using neural networks, which are better at learning  unknown distributions.
Along the direction of the spatio-temporal Hawkes process, \cite{du2016recurrent} has considered spatial attributes by discretizing space and considering them as marks, which is unable to model spatio-temporal intensity function. \cite{okawa2019deep,zhou2021neural} has proposed spatio-temporal intensity as a mixture of kernels. \cite{chen2020neural} has proposed Neural ODE for modeling spatio-temporal point processes.
Another work \cite{ilhan2020modeling} has applied random Fourier features based transformations to represent kernel operations in spatio-temporal Hawkes process. A recent work~\cite{zhou2021neural} has introduced deep spatio-temporal point process where they integrate spatio-temporal point process with deep learning by modeling space-time intensity function as mixture of kernels where intensity is inferred using variational inference. 


Bayesian Neural Networks (BNNs) \cite{neal2012bayesian,gal2016uncertainty} are widely used framework to find uncertainty estimates for deep models. However, one has to resort to approximate inference techniques~\cite{graves2011practical,neal2012bayesian} due to intractable posterior in BNNs. An alternative solution is to incorporate uncertainty directly into the loss function~\cite{lakshminarayanan2017simple,pearce2018high}. 
Monte Carlo Dropout proposed by~\cite{gal2016dropout,gal2016theoretically} proved that dropout regularization~\cite{srivastava2014dropout} can act as approximation for Bayesian inference over the weights of neural networks. 


There are few works~\cite{zhang2020variational,zhang2019efficient,wang2020uncertainty} where uncertainty is captured over the parameters of parametric hawkes process.  In~\cite{chapfuwa2020calibration} authors have proposed an adversarial non-parametric model that accounts for calibration and uncertainty for time to event and uncertainty prediction.
However, neural point processes are a widely adopted model for event modeling due to its theoretical and practical effectiveness in various applications. Therefore, the goal of this paper is to augment the capabilities of the neural point process to predict future events  with the uncertainty estimation capability. To achieve this, we  propose a novel approach  which  combines neural point process and  Monte Carlo dropout for performing uncertainty estimation for event-time prediction.

\section{Bayesian Neural Hawkes Process}
We consider the input sequence $S = \{t_i\}_{i=1}^N$ in the observation interval $[0, T]$ and inter-event time interval as $\tau_i = t_i - t_{i-1}$.
Our goal is to predict the time of occurrence of next events along with uncertainty estimates over the predicted time. 
We consider a neural Hawkes process (NHP)~\cite{omi2019fully}, which consider  a combination of recurrent neural network and feed-forward neural network to model the likelihood associated with the sequence of events $\{t_i\}_{i=1}^N$ and predict the future events. We propose  Bayesian Neural Hawkes process by developing MC dropout jointly over recurrent and feed-forward neural networks in an NHP. 

\subsection{Neural Hawkes Process}
\label{Section:NHP}
 A major characteristic of the Hawkes process~\cite{hawkes1971spectra} is   the conditional intensity function which conditions the next event occurrence based on the history of events. Following recent advances in Hawkes process~\cite{du2016recurrent,omi2019fully}, neural Hawkes processes were proposed which model the intensity function as a nonlinear function of history  using a neural network. In particular \cite{omi2019fully} uses   a combination of recurrent neural network and feedforward neural network to model the intensity function. This  allows the intensity function to take any functional form depending on  data and help in better generalization performance.  
We represent history by using hidden representations generated by recurrent neural networks (RNNs) at each time step. The hidden representation $\boldsymbol{h}_i$ at time $t_i$ is obtained as
\begin{equation}
\boldsymbol{h}_i = RNN(\tau_i, \boldsymbol{h}_{i-1}; W_r ) = \sigma(\tau_i V_r +  \boldsymbol{h}_{i-1} U_r + b_r)
\label{eq:rnn}
\end{equation}
where $W_r$ represents the parameters associated with RNN such as  input weight matrix $V^r$, recurrent weight matrix $U_r$, and and bias $b_r$.  $\boldsymbol{h}_i$ is obtained by repeated  application of the RNN block on a sequence formed from  previous $M$ inter-arrival times.
This is used as input to a feedforward neural network to compute the intensity function (hazard function) and consequently the cumulative hazard function for computing the likelihood of  event occurrences.

In the proposed model, we consider the following input to the feed-forward neural network \Romannum{1}) the hidden representation generated from RNN,  \Romannum{2}) time of occurrence of the event and \Romannum{3}) elapsed time from the most recent event.  To better capture the uncertainty in predicting points in future time, we considered  the time at which the intensity function needs to be evaluated. 
We model the conditional intensity as a function of the elapsed time from the most recent event,  and is represented  as $\lambda(t-t_i|\boldsymbol{h}_i, t)$,
where $\lambda(\cdot)$ is a non-negative function referred to as a hazard function.
 Therefore, we define cumulative hazard function in terms of inter-event interval 
 \begin{equation}
 \tau = t - t_{i} = \Phi(\tau|\boldsymbol{h}_i, t) = \int_{0}^{\tau}\lambda(s|\boldsymbol{h}_i, t)ds
 \end{equation}
Cumulative hazard function is modeled using a  feed-forward  neural network (FNN)
\begin{equation}
\Phi(\tau|\boldsymbol{h}_i, t ) = FNN(\tau, h_i, t; W_t).
\end{equation}
However, we need to fulfill two properties of cumulative hazard function. Firstly, it has to be a monotonically increasing function of $\tau$ and secondly, it has to be positive valued. We achieve these by maintaining positive weights and positive activation functions in the neural network~\cite{chilinski2020neural,omi2019fully}. The hazard function itself can be then obtained by differentiating the cumulative hazard function with respect to $\tau$ as 
\begin{equation}
\lambda(\tau|\boldsymbol{h}_i, t) = \frac{\partial}{\partial \tau}\Phi(\tau|\boldsymbol{h}_i, t)
\end{equation}
The log-likelihood of observing event times is defined as follows using the cumulative hazard function:
\begin{equation}
\begin{split}
\label{eq:log_lik_nhp}
& \log p(\{t_i\}_{i=1}^N ; W) = \sum_{i=1}^N \log p(t_i|\mathcal{H}_i;W) = 
 \sum_{i=1}^N \big( \log (\frac{ \partial}{\partial \tau} \Phi(\tau_i |\boldsymbol{h}_{i-1}, t_{i}; W) )
 - \Phi(\tau_i|\boldsymbol{h}_{i-1}, t_{i}; W) \big) 
 \end{split}
\end{equation}
where $\tau_i = t_{i}-t_{i-1}$ and $W = \{W_r, W_t\}$ represents the combined weights associated with  RNN and  FNN.  In NHP, the weights of the networks are learnt by maximizing the likelihood given by (\ref{eq:log_lik_nhp}). The gradient of the log-likelihood function is calculated using backpropagation. 


\begin{figure}
    \centering
    \includegraphics[scale=0.20]{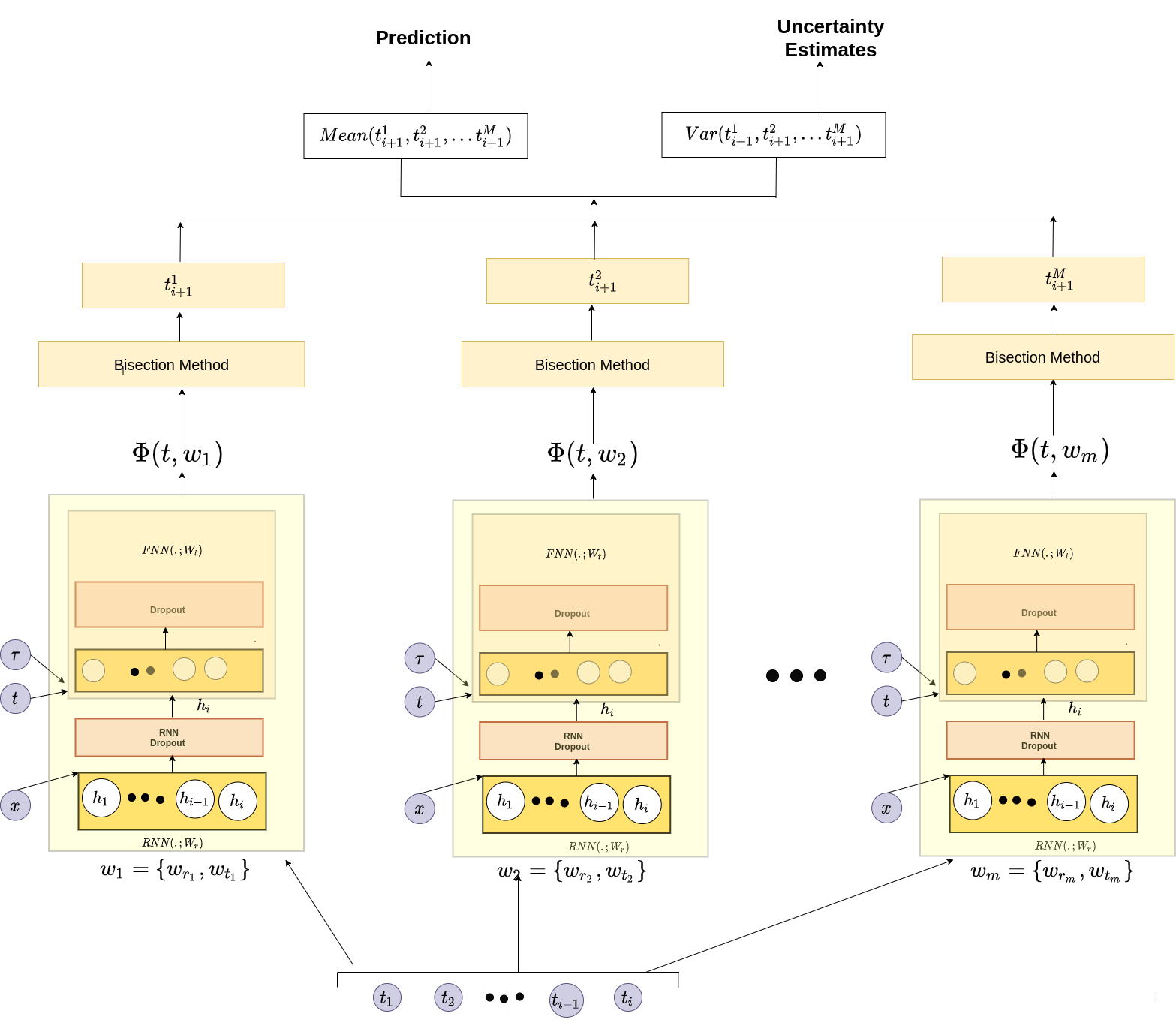}
    \caption{Framework for the proposed approach. Input: Time of occurrence associated with sequence of events. Output: time of occurrence of future events along with the uncertainty estimate.}
    \label{fig:framework_uncertainty}
\end{figure}

\subsection{Monte Carlo Dropout for Neural Hawkes Process}
\label{sect:bnhp}
In this section, we discuss our proposed framework  \textit{Bayesian Neural Hawkes Process (BNHP)} (shown in Figure \ref{fig:framework_uncertainty}),  for estimating uncertainty over time of occurrence of  events.
We propose a simple and practical approach based on Monte Carlo (MC) Dropout~\cite{gal2016dropout}] to model uncertainty in NHP.
We develop MC Dropout for the NHP model which consists of two kinds of networks -  RNN processing the input sequence and  FNN producing the cumulative hazard function. 


To model uncertainty, we follow a  Bayesian  approach, where we treat the weight matrices as random variables and  put a prior distribution such as a standard Gaussian over them, i.e. $p(W)  = \mathcal{N}(W;0,I)$. 
A Bayesian model will learn a posterior distribution over the parameters, instead of point estimates which allows them to capture uncertainty in the model. 
The  likelihood of observing the event times given the weight parameters is defined by \eqref{eq:log_lik_nhp}. 
We find the posterior distribution over the parameters  $p(W | \{t_i\}_{i=1}^N )$ given the observed sequence of events  using the Bayes Theorem, 
 \begin{equation}
     \begin{aligned}
         p(W \mid \{t_i\}_{i=1}^N )=\frac{p(\{t_i\}_{i=1}^N \mid W ) p(W)}{p(\{t_i\}_{i=1}^N)}.
     \end{aligned}
 \end{equation}
However, the posterior computation is intractable due to non-linear functions modelled using neural networks. Hence, we find an approximate distribution $q(W)$ to the posterior  $p(W | \{t_i\}_{i=1}^N)$ using variational inference (VI)~\cite{blei17}.  The parameters of  $q(W)$ is obtained by minimizing the  variational lower bound (ELBO) involving Kullback-Leibler (KL) divergence. 
\begin{eqnarray}
- \sum_{i=1}^N \int q(W)\, \log p(t_i|\mathcal{H}_i;W) \,dW  + KL(q(W) \parallel p(W)) \nonumber
\label{eqn:elbo}
\end{eqnarray}
Monte-Carlo (MC) dropout~\cite{gal2016dropout} assumes a particular form for the variational approximation which makes the ELBO computation equivalent to training the neural networks with dropout regularization.
It has been shown that using  VI with this specific form for the variational approximation  is equivalent to applying dropout during training and testing~\cite{gal2016dropout}. 

A challenge with the neural Hawkes process in applying MC dropout is that, we have two kinds of neural networks  with two types of weight parameters:  $W_t$ associated with the FNN and $W_r$ associated with the RNN. Consequently, we use different variational approximations over each of the parameter types and thus different dropout regularizations as described below. We assume the variational distribution $q(W)$ can be decomposed as product of the variational distributions $q(W_t)$ and $q(W_r)$. We assume the following form for the variational distribution $q(W_t) = \prod_{l=1}^L q(W_t^l)$, where  a sample  $\hat{W}_t^l$  associated with layer $l$   is obtained from   $q(W_t^l)$ as
\begin{eqnarray}
\label{eq:vardistr}
 \hat{W}_t^l \sim \bar{W}_t^l \cdot diag([\mathbf{z}^l]) \quad \, \quad \mathbf{z}^l \sim Bernoulli(p)  
\end{eqnarray}
where $p$ is the drop probability and $\bar{W}_t^l$ are variational parameters (also weight matrix) to be estimated from the ELBO. Due to binary valued $\mathbf{z}^l$,  $\hat{W}^l$ sampled from $q(W^l)$ will have some of the columns of the weight matrix set to zero aiding in dropout regularization.

For the weights $W_r$ in the RNN ,  we use  a variant of the standard MC-Dropout to perform learning and inference. First we assume the variational distribution $q(W_r)$ to factorize over the weight matrices $V^r$, $U_r$, and $b_r$, 
\begin{equation}
q(W_r) = q(V_r)q(U_r)q(b_r)
\end{equation}
For each of these weight matrices, we assume the variational distribution to factorize over the rows, 
\begin{equation}
q(V_r) = \prod_{k=1}^K q(V_{rk}) 
\end{equation}
Each row in the weight matrix follows an approximate distribution :
\begin{equation}
    q(V_{rk}) = p \mathcal{N}(V_{rk};0,\sigma_r^2 I) + (1-p) \mathcal{N}(V_{rk};\bar{V}_{rk}, \sigma_r^2 I)
    \label{eqn:rnnvar}
\end{equation}
where $p$ is the drop probability, $\bar{V}_{rk}$ is the weight vector (also variational parameters) and $\sigma_r^2$ is the variance.  The distribution \eqref{eqn:rnnvar} is similar to the sampling process in \eqref{eq:vardistr} when the variance is small. We sample the weight matrices $\hat{W}_r = \{ \hat{V}_r, \hat{U}_r, \hat{b}_r\}$ associated with RNN from the variational distribution defined by \eqref{eqn:rnnvar}. This results in the masking of random rows of the weight matrices. In the case of RNN, the same masked weight matrices are used for processing all the elements in the input sequence~\cite{gal2016theoretically}. But, different sequences use different masked weights. 

For obtaining  the distribution over the inter-arrival time $\tau_i$, we first obtain hidden representation $\mathbf{h}_{i-1}$  using  an RNN with sampled weight matrices $\hat{W}_{ri}$. This is then fed through FNN along with other inputs and sampled weight matrix $\hat{W}_{ti}$ to obtain the cumulative hazard function $\Phi(\tau_i|\boldsymbol{h}_{i-1}, t_i, \hat{W}_i)$. 
Consequently, the variational lower bound (ELBO) is written as 
\begin{eqnarray}
& L_{VI}(\bar{W})
= - \sum_{i=1}^N \log p(t_i|\mathcal{H}_i;\hat{W}_i)  + \lambda \parallel \bar{W} \parallel^{2} \nonumber \\
\label{eqn:elbodrop}
&=- \sum_{i=1}^N \big( \log (\frac{ \partial}{\partial \tau} \Phi(\tau_i |\boldsymbol{h}_{i-1}, t_{i}; \hat{W}_i) )
 - \Phi(\tau_i|\boldsymbol{h}_{i-1}, t_{i}; \hat{W}_i) \big) \nonumber
  + \lambda ( \parallel \bar{W}_t \parallel^{2} +  \parallel \bar{V}_r \parallel^{2} +   \parallel \bar{U}_r \parallel^{2}  +  \parallel \bar{b}_r \parallel^{2}) 
\end{eqnarray}
where $\bar{W}  = \{\bar{W}_t, \bar{V}_r, \bar{U}_r, \bar{b}_r\}$ represents all the weight matrices (variational parameters) that needs to be learnt by minimizing the objective function  \eqref{eqn:elbodrop}. We can observe that this is equivalent to learning  weight parameters in a neural Hawkes process by minimizing the negative log-likelihood \eqref{eq:log_lik_nhp}, but with dropout regularization and $\ell_2$ norm regularization (obtained by simplifying the KL term) over the weight matrices.

\subsection{Prediction and Uncertainty Estimation}
\label{sec:pred_uncertainty}
For prediction, Bayesian models make use of the full posterior distribution over parameters rather than a point estimate of the parameters. This allows them to make robust predictions and model uncertainty over the predictions. After learning the variational parameters $\bar{W}$, we use the approximate  posterior $q(W)$ to make predictions.  The probability of predicting the time of next event given the history of previous event times with last event at $t_N$ is computed as 
\begin{equation}
p(t_*|\mathcal{H}) = \int p(t_*|\mathcal{H};W) q(W) dW = \frac{1}{S} \sum_{s=1}^S p(t_*|\mathcal{H};\hat{W}_s)  \nonumber
\end{equation}
where we approximated the integral with a Monte-Carlo approximation~\cite{neal2012bayesian} with $S$ samples from $q(W)$ ($\hat{W}_s \sim q(W) $) and $\tau_* = t_* - t_N$. We can observe that sampling from $q(W)$ leads to  performing dropout at the test time and the final prediction is obtained by averaging the prediction over multiple dropout architectures.  Here, uncertainty in the weights induces model uncertainty for event time prediction.

Monte-Carlo dropout on NHP provides a  feasible and practical technique to model uncertainty in event prediction. MC dropout is equivalent to performing $M$ forward passes through the NHP with  stochastic dropouts at each layer but considering the specific nature of the network (recurrent or feedforward).  
Final prediction is done by taking an average prediction from the $M$ network architectures formed out of dropout in the forward pass. This can be seen as equivalent to ensemble of NHPs which could also lead to improved predictive performance. 
For each network, we use the bisection method~\cite{omi2019fully} to predict the time of the next event. Bisection method provides  the median  $t_*$ of the predictive distribution over next event time using  the relation $\Phi(t_* - t_N|\boldsymbol{h}_N, t_*; \hat{W})=\log(2)$. We obtain $S$ median event times, with each median event time $t_{*_s}$ obtained using a network with sampled  weight $\hat{W}_s$.   The mean event time is then found by averaging  $S$ times obtained from different sampled weights, $t_* = \frac{1}{S} \sum_{s=1}^S t_{*_s}$.
The variance of predicted time is obtained as 
\begin{equation}
Var(t_*) = {\sigma_*}^2 = \frac{1}{S} \sum_{s=1}^S (t_{*_s} - t_*)^2
\end{equation}
We also define lower and upper bound of predicted time as 
\begin{equation}\label{eq:bound} 
{t}_{*}^L = {t}_* - k\sigma_* ; \quad
{t}_{*}^U = {t}_* + k\sigma_* 
\end{equation}
Algorithm~\ref{alg:algorithm1} provides the details of the prediction process,  where $\hat{W}_s$ denotes a weight vector sample obtained using dropout at test time. 

\begin{algorithm}
\begin{algorithmic}[1]
\State \textbf{Input:} $\{t_1, t_2, t_3, \ldots, t_p\}$, number of MC samples $S$, parameters $W$ and neural HP network $f^W(\cdot)$ which consider previous $M$ inter-arrival times. 
\State\textbf{Output:} Time of occurrence of future predictions $\hat{t}_{p+1}, \hat{t}_{p+2}, \ldots, \hat{t}_{n} $ and uncertainty intervals as $\{\hat{t}_{p+1}^L$, $\hat{t}_{p+1}^U\}$, $\{\hat{t}_{p+2}^L$, $\hat{t}_{p+2}^U\}$ and $\{\hat{t}_{n}^L$, $\hat{t}_{n}^U\}$
\State Train the model to learn $\bar{W}$ from the train data $\{t_1, t_2, t_3, \ldots, t_p\}$ by minimizing the loss \eqref{eqn:elbodrop}. 
\For{$i\gets p+1,n$}
\For{$s \gets 1, S$}
\State $\Phi_i^{s}(\tau) \gets f^{(\hat{W}_s)}(\{\tau_k,t_k\}_{k=i-M}^{i-1})$
\State $\lambda_i^{s}(\tau) \gets \frac{\partial}{\partial \tau} \Phi_i^{s}(\tau)$ 
\State $\bar{t}_{i}^{(s)} \gets \textsc{Bisection}(\Phi_i^{s}(\tau))$ 
\EndFor
\State $\hat{t}_{p} \gets \frac{1}{S} \sum_{s=1}^S \bar{t}_{i}^{(s)} $ 
\State $\sigma_p^2 \gets \frac{1}{S} \sum_{s=1}^S (\bar{t}_{p}^{(s)} - \hat{t}_{p})^2$ 
\State $\hat{t}_{p}^L = \hat{t}_{p} - k*\sigma_p$,  $\hat{t}_{p}^U = \hat{t}_{p} + k*\sigma_p$
\EndFor
\end{algorithmic}
\caption{Proposed Approach}
\label{alg:algorithm1}
\end{algorithm}

\subsection{Bayesian Neural Hawkes Process for Spatio-Temporal Modeling (ST-BNHP)}

We extend the Bayesian Neural Hawkes Process developed for modeling uncertainty over time to modeling uncertainty over space. For applications involving spatio-temporal data such as traffic  flow prediction, and human mobility prediction it is useful to obtain uncertainty over spatial points apart from the temporal uncertainty. 
Following \cite{valkeila2008introduction}, we model conditional intensity function for modelling the spatio-temporal data as $\lambda(t,x) = \lambda(t) p(\mathbf{x} | t)$,
where $\mathbf{x} \in \mathcal{R}^D$ represents a spatial  point or location and $p( \mathbf{x}| t)$ is the conditional spatial density of location $\mathbf{x}$ at t given $H_t$. 

Since many mobility patterns involve movement in close-by regions, we model the conditional spatial density $p(\mathbf{x}_i|t_i)$ as a Gaussian conditioned on the history, i.e.  all the  past events, times and locations ($(t_{i-}, \mathbf{x}_{i-})$). Thus,  
\begin{equation}
\begin{split}
\label{eqn:Gaussian}
  p(\mathbf{x}_i | t_{i-}, \mathbf{x}_{i-}) &= \mathcal{N}(\mathbf{x}_i ; \mu(t_{i-}, \mathbf{x}_{i-}),
  \sigma(t_{i-}, \mathbf{x}_{i-}))
    \end{split}
\end{equation}
In the proposed model, we consider  parameters of the Gaussian  $\mu(\cdot)$ and $\sigma(\cdot)$ as functions of past event times and locations and  are modeled using a feed-forward neural network.
\[
     [\mu(t_{i-}, \mathbf{x}_{i-}), \sigma(t_{i-}, \mathbf{x}_{i-}) ]= FNN(\boldsymbol{h}_i, \mathbf{x}_{i-};W_l)
\]
where $\boldsymbol{h}_i$ is the hidden representation obtained from the RNN model  with the sequence of the past event times as input. 
 
The log likelihood for the  spatio-temporal neural Hawkes model can be written as $\sum_{i=1}^N  LL(t_i,\mathbf{x}_i; W,W_l)$, where 
\begin{equation}
\label{eq:log-likelihood-spatial}
LL(t_i,\mathbf{x}_i; W,W_l) = \log p(t_i|\mathcal{H}_i;W) + \log p(\mathbf{x}_i| t_{i-}, x_{i-}; W_l). \nonumber
\end{equation}
To obtain spatio-temporal Bayesian neural Hawkes (ST-BNHP) model, we consider a distribution over $W_l$ and  follow a similar approach as in Section \ref{sect:bnhp}, where a variational distribution  $q(W_l)$ over $W_l$  is defined as in \eqref{eq:vardistr}. The resulting ELBO considers the spatial likelihood and can be written as
\begin{equation}
\label{eq:log-likelihood-spatial-montecarlo}
\begin{split}
 & - \sum_{i=1}^N LL(t_i,\mathbf{x}_i;\hat{W}_i,\hat{W}_{li}) + \lambda (\parallel \bar{W} \parallel^{2} + \parallel \bar{W}_l \parallel^{2} )     \end{split}
\end{equation}
where $\hat{W}_{li} \sim q(W_l)$ and $\bar{W}_l$ is the variational parameter (weight matrix). The  parameters  are learnt by minimizing the lower bound \eqref{eq:log-likelihood-spatial-montecarlo} and is equivalent to learning with dropout regularization. Prediction is done as discussed in  Section \ref{sect:bnhp}, where one considers multiple dropout architectures (corresponding to different samples of weight matrices), and  taking an average over them.  For each dropout architecture, we predict the time of occurrence of the next event using the bisection method and the location of of the next event by sampling from the conditional Gaussian $\mathcal{N}(\mathbf{x}_* ; \mu(t_{N-}, \mathbf{x}_{N-}),
  \sigma(t_{N-}, \mathbf{x}_{N-}))$. 

\begin{table*}
\centering
\tabcolsep=0.05cm
\begin{tabular}{c|cccc|cccc|cccc|cccc} 
\hline
\begin{tabular}[c]{@{}c@{}}\\\end{tabular} &      & \multicolumn{4}{c}{Sim-Poisson}        & \multicolumn{4}{c}{Sim-Hawkes}        & \multicolumn{4}{c}{Crime}             & \multicolumn{3}{c}{Music}       \\ 
\hline
                           \textbf{Metric}                & SHP  & NHP~ & EH    & BNHP           & SHP   & NHP & EH    & BNHP           & SHP   & NHP & EH    & BNHP           & SHP   & NHP & EH    & BNHP             \\ 
\hline
\textbf{MNLL}                              & 1.21 & 1.36  & *     & \textbf{0.196} & 1.08  & 1.26 & *     & \textbf{0.166} & 1.809 & 1.42 & *     & \textbf{0.279} & 1.239 & 0.97 & *     & \textbf{-0.322}  \\
\textbf{MAE}~                              & 0.81 & 0.76  & 0.172 & \textbf{0.078} & 0.382 & 0.83 & 0.234 & \textbf{0.156} & 0.903 & 0.82 & 0.242 & \textbf{0.158} & 0.997 & 0.58 & 0.104 & \textbf{0.060}   \\
\textbf{PIC@1}                             & 0.0  & 0.0     & 0.019 & \textbf{0.946} & 0.0   & 0.0    & 0.012 & \textbf{0.725} & 0.0   & 0.0    & 0.002 & \textbf{0.586} & 0.0   & 0.0    & 0.265 & \textbf{0.836}   \\
\textbf{PIC@2}                             & 0.0  & 0.0     & 0.031 & \textbf{0.996} & 0.0   & 0.0    & 0.029 & \textbf{0.976} & 0.0   & 0.0    & 0.005 & \textbf{0.856} & 0.0   & 0.0    & 0.312 & \textbf{0.963}   \\
\textbf{PIC@5}                             & 0.0  & 0.0     & 0.061 & \textbf{1.0}   & 0.0   & 0.0    & 0.058 & \textbf{0.999} & 0.0   & 0.0    & 0.015 & \textbf{0.997} & 0.0   & 0.0    & 0.458 & \textbf{0.996}   \\
\hline
\end{tabular}
\caption{Comparison of the proposed approach (BNHP) against the baselines (Please note that EH uses least square loss and hence MNLL for EH is not comparable to our results, hence MNLL for EH is marked with '*'. ) }
\label{Tab:experiments}
\end{table*}
    
\begin{table*}
\centering
\begin{tabular}{cccccccccc} 
\hline
Dataset                           & Method    & MNLL   & \begin{tabular}[c]{@{}c@{}}MNLL\\Temporal\end{tabular} & \begin{tabular}[c]{@{}c@{}}MNLL\\Spatial\end{tabular} & \begin{tabular}[c]{@{}c@{}}MAE\\time\end{tabular} & \begin{tabular}[c]{@{}c@{}}MAE\\lat\end{tabular} & \begin{tabular}[c]{@{}c@{}}MAE\\long\end{tabular} & \begin{tabular}[c]{@{}c@{}}PIC\\time\end{tabular} & \begin{tabular}[c]{@{}c@{}}PIC\\spatial\end{tabular}  \\ 
\hline
\multirow{1}{*}{Foursquare} & ST-HomogPoisson & 8.204  & -                                                      & -                                                     & 4.229                                             & 1.23                                             & 1.824                                             & 0.026                                             & 0.025                                                 \\
                                     & ST-NHP & 5.366  & 1.562                                                  & 3.804                                                 & 2.048                                             & 0.889                                           & 0.798                                             & 0.0                                             & 0.387                                                         \\ 
                                 & ST-BNHP & \textbf{2.817}  & \textbf{0.477}                                                  & \textbf{2.340}                                                 & \textbf{1.122}                                             & \textbf{0.754 }                                           & \textbf{0.760}                                             & \textbf{0.877}                                             & \textbf{0.649}                                                         \\ 

\hline
\multirow{2}{*}{Taxi}  & ST-HomogPoisson       & 13.711 & -                                                      & -                                                     & 2.472                                             & 4.386                                            & 1.527                                             & 0.041                                             & 0.056                                                          \\
 & ST-NHP & 3.812  & 1.476                                                  & 2.335                                                 & 1.501                                             & 0.639                                           & 0.581                                            & 0.0                                           & 0.546                                                         \\ 
                                 & ST-BNHP       & \textbf{2.392}  & \textbf{0.054}                                                  & \textbf{2.338}                                                 & \textbf{0.133}                                             & \textbf{0.589}                                            & \textbf{0.536}                                             & \textbf{0.386}                                             & \textbf{0.721}                                                 \\
\hline
\end{tabular}
\caption{Comparison of the proposed approach (ST-BNHP) against the baselines for spatio-temporal modeling }
\label{Tab:st-experiments}
\end{table*}

  \begin{table}[t]
\centering
\caption{Dataset Statistics}
\label{Tab:dataset_stats}
\begin{tabular}{|c|c|} 
\hline
Dataset     & \# events  \\ 
\hline
Sim-Poisson & 80,000     \\ 
\hline
Sim-Hawkes  & 80,000     \\ 
\hline
Crime       & 217,662    \\ 
\hline
Music       & 112,610    \\ 
\hline
NYC         & 200,000    \\ 
\hline
Foursquare  & 95,702     \\
\hline
\end{tabular}
\end{table}
\section{Experiments and Results}
\paragraph{Dataset Details}
We extensively perform experiments on six datasets including two synthetic and four real-world datasets. Dataset statistics are mentioned in Table \ref{Tab:dataset_stats}.
\begin{itemize}

 \item \textbf{Simulated Poisson (Sim-Poisson):}   We simulate a homogeneous poisson process with conditional intensity $\lambda$ = 1.
  \item  \textbf{Simulated Hawkes (Sim-Hawkes):} We use the Hawkes process, in which the kernel function is given by the sum of multiple exponential functions. The conditional intensity function is given by $\lambda(t) = \mu + \sum_{t_i < t} \sum_{j=1}^2 \alpha_j \beta_j \exp{(- \beta_j(t - t_i))}$. We have used $\mu = 0.05 $, $\alpha_1 =  0.4$, $\alpha_2 = 0.4$, $\beta_1 = 1.0$, $\beta_1 = 20.0$.
  \item   \textbf{Crime:} This dataset contains the records of the police department calls for service in San Francisco~\footnote{\url{https://catalog.data.gov/dataset/police-calls-for-service}}. Each record contains the crime and timestamp associated with the call along with other information.  We have selected ten such crimes as \textit{Burgalary, Drugs, Fraud, Injury, Accident, Intoxicated Person, Mentally disturbed, Person with knife, Stolen Vehicle, Suicide Attempt, Threats/Harrassment} and their associated timestamps. Each sequence corresponds to a crime.
  \item  \textbf{Music:} This dataset contains the history of music listening of users at \textit{lastfm}~\footnote{\url{https://www.dtic.upf.edu/~ocelma/MusicRecommendationDataset/lastfm-1K.html}}. We consider 20 sequences from 20 most active users in the month of January, 2009. Each sequence corresponds to one user.
  \item  \textbf{New York Taxi:} The dataset contains trip records for taxis in New York City. Each pick-up record is considered as an event. This datset contains spatial attributes in the form of latitude-longitude pair along with time of occurrence of events. We consider 100,000 events for each user and a sequence is constructed for each vendor. 
   \item  \textbf{Foursquare:} This is a location-based social network which contains information about check-ins of users. We have considered top 200 users during the period April 2012 to February, 2013. Each sequence corresponds to each user. This dataset also contains spatial attributes as latitude-longitude along with time of occurrence of events. 
    \end{itemize}
We have used the New York Taxi and Foursquare dataset for spatio-temporal modeling.

\paragraph{Evaluation metrics}

We consider these metrics for evaluation of our results 
\begin{itemize}
 \item  \textbf{Mean Negative Log Likelihood (MNLL): } Log-likelihood considers the probability  of predicting the actual observations, and we expect a good model to have a lower MNLL score.  
    \item  \textbf{Mean Absolute Error for prediction (MAE):} The measure computes purely the absolute error in predicting the time without considering the probability. A model with low MAE will reflect a better model.
    \item  \textbf{Prediction Interval Coverage (PIC):} It represents the number of times an actual event occurs within the estimated interval. Hence, higher PIC will represent a better model.
    \item  \textbf{Prediction Interval Length (PIL):} It represents the length of difference between upper bound and lower bound of the predicted interval (Refer \eqref{eq:bound}). We report the  mean and variance of interval length ($2*k*\sigma$).  Higher PIL signifies more uncertainty associated with the event.
  \end{itemize} 
  These evaluation metrics are tweaked for spatio-temporal modeling as well. For example, spatial MNLL represents conditional spatial density. MAE and PIL are calculated separately for latitude and longitude. Moreover, we modulate PIC in spatial manner by considering an actual event to occur within the estimated latitude and longitude bounds.

\paragraph{Baselines}
We compare our experimental setup with the following baselines. 
\begin{itemize}

\item  \textbf{NHP:} Neural Hawkes process (NHP)~\cite{omi2019fully} serves as a baseline for evaluating the performance of predicted time of occurrence of event. 
    \item  \textbf{SHP:} We consider the standard Hawkes Process with exponential kernel as another baseline. NHP and SHP  model probability distribution over event times  $p(\tau | h_i) $, though they do not model epistemic uncertainty like BNHP. 
   \item  \textbf{Ensemble Hawkes (EH):} We propose a new baseline capable of modelling epistemic uncertainty - an ensemble of parametric HP using exponential kernels  with different hyper-parameters. 
    To avoid convergence issues, we use least-square loss \cite{xu2018benefits} for training EH. Due to this, MNLL of EH won't be comparable with other methods which use survival likelihood. Each ensemble consists of 10 models with  decays ranging between 0.001 to 0.1.
     \item  \textbf{ST-HomogPoisson:} The intensity is assumed to be constant over space and time. This is used as a baseline against ST-BNHP. 
     \item  \textbf{ST-NHP:} is the NHP augmented with the Gaussian spatial conditional density defined in \eqref{eqn:Gaussian}.
\end{itemize}

\paragraph{Implementation Details}
Herein, we present training details for all our experiments.  We split our dataset into training, validation and test set as 70-10-20. The split is made across all sequences. Therefore, the first 70\%, next 10\% and last 20\% of the events in each sequence are used for training, validation and testing respectively. The best hyperparameters are selected using the validation set. We consider a  recurrent neural network with one layer and 64 units and  5-layer feed-forward neural network with 16 units in each layer. Recurrent Neural Network is associated with a dropout at input level as well as recurrent layer level. Along with that, there is a dropout associated with feedforward neural network to represent cumulative hazard function. Also, we use \textit{Tanh} activation function. We use Adam optimizer with learning rate, $\beta_1$ and $\beta_2$ as 0.0001, 0.90 and 0.99 along with L2 regularization. We have used 50 different architectures using Monte Carlo Dropout. The results are reported after averaging the results after running the model three times. We perform 1-step lookahead prediction where we use actual time of occurrence of events as past events for the historical information to predict future event. We have perfomed all the experiments on Intel(R) Xeon(R) Silver 4208 CPU @ 2.10GHz, GeForce RTX 2080 Ti GPU and 128 GB RAM. We have implemented our code in Tensorflow 1.5.0 \cite{tensorflow2015-whitepaper}.

\subsection{Results}
We perform evaluation of the proposed methodology across quantitative and qualitative dimensions. We compare our results with the baselines and are reported in Table~\ref{Tab:experiments}. We can observe that the proposed methodology is performing well in terms of time of prediction of event as well as uncertainty estimation for the event. The proposed method BNHP achieves significantly better MNLL and MAE  as compared to NHP, SHP and ensemble of Hawkes Process for all the datasets.  A better generalization performance can be attributed to the ensemble of NHP architectures obtained through dropout at test time. In addition to improved predictive performance, our approach also provides better  uncertainty estimates. We quantify the uncertainty interval on the basis of how well the predicted intervals capture the actual occurrence of events. We report \textit{PIC@k}   in Table~\ref{Tab:experiments} for different values of  k ($k=\{1,2,3\}$) in computing the upper bound and lower bound of the prediction interval(Refer \eqref{eq:bound}). SHP and NHP provide very small interval lengths and the actual points will not lie in this small interval, hence their PIC values are close to $0$. So, we compare the PIC of the proposed approach  with an ensemble of Hawkes Processes. We can observe that we predict better PIC values compared to EH. We augment the proposed model to spatio-temporal event modeling as well where we predict uncertainty associated with time and location of the event. We report the results in Table \ref{Tab:st-experiments}. The proposed extension (ST-BNHP) performs better than both the baselines for both the datasets in terms of MNLL and MAE. We can observe that \textit{PIC@1} values are also better for predicted time and location (in terms of latitude and longitude) of occurrence of an event. One could argue that too wide intervals may yield higher \textit{PIC}, but such intervals may be of no use. However, if the predicted intervals are too high, the algorithm will output higher MNLL score (or smaller log-likelihood). On the contrary, if the interval length is too small, then the number of events captured within the predicted interval will also be less, hence yielding a low PIC. Ideally, the length of the uncertainty interval should  be large enough to capture points well and small enough to yield low MNLL. We observe that the proposed BNHP model gives lowest MNLL and highest PIC. Also, MAE scores are also better for the real-world datasets. Our method achieves significantly  better results in terms of MNLL, MAE and PIC both on synthetic and real world data sets, establishing it as an effective technique for spatio-temporal modelling.

\subsection{Analysis}
  \begin{figure}[t]
  \centering
   \subfloat[Sim-Poisson Dataset] {\includegraphics[scale=0.28]{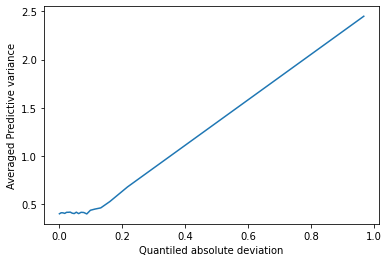} \label{fig:quantile_sim}} \quad
   \subfloat[Crime Dataset]{\includegraphics[scale=0.28]{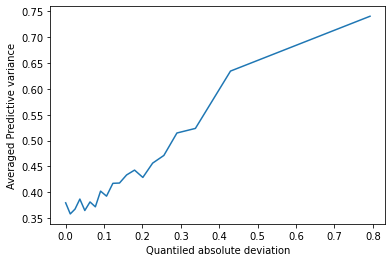}  \label{fig:quantile_crime}}
     \caption{Predictive variance  against quantiles of absolute deviation. }
      \label{fig:quantile_plots}
   \vspace{-4mm}
  \end{figure}
  
  \begin{table}[t]
  \vspace{-2mm}
\centering
\begin{tabular}{c|c|c|c|c} 
\hline
\textbf{Dataset}     & Sim-Poisson & Sim-Hawkes &  Music & Crime  \\ 
\hline
\textbf{correlation} & 0.537  & 0.324   & 0.336      & 0.107   \\                                
\hline
\vspace{-5mm}
\end{tabular}
\caption{Spearman Coefficient between MAE and PIL (positive value indicates positive correlation) }
\label{Tab:correlation}
\end{table}
\paragraph{Relevance of Predictive Uncertainty}
Evaluating uncertainty for an event data is a difficult task. We expect a model to be more confident for correct predictions and should exhibit smaller uncertainty. Consequently, it should assign lower PIL for correct predictions and vice-versa. To acknowledge this behavior, we plot Quantile Plots of absolute deviation against the average variance predicted by the model. For this, we consider quantile $q$ of the absolute deviation (AD) of predicted time and actual time $| \hat{t}_i - t_i |$. Then we average prediction interval length (Avg-PIL), of all the events with absolute deviation less than the quantile value. 
\[
\text{Avg-PIL} = \{ \text{Average}(2*\sigma_i) | \quad \forall |t_i - \hat{t}_i| <= \text{q-th quantile}\}
\]
We can observe the quantile plots of averaged predictive variance against the absolute deviation of the predicted time for simulated Poisson and Crime dataset in the Figure \ref{fig:quantile_plots}. An increasing line suggests that as absolute deviation increases, predictive uncertainty also increases, hence supporting our hypotheses.
We also validate our hypothesis using correlation between predictive variance and AD. We calculate Spearman coefficient between AD and PIL for all the datasets  and report in Table~\ref{Tab:correlation}. Here, we can observe that the Spearman coefficient is positive for all datasets, supporting our hypothesis.

      \begin{figure}[t]
  \centering
  \caption{Correlation between MAE and PIL}
 \label{fig:mae_corr2}
  \subfloat[PIL] {\includegraphics[scale=0.24]{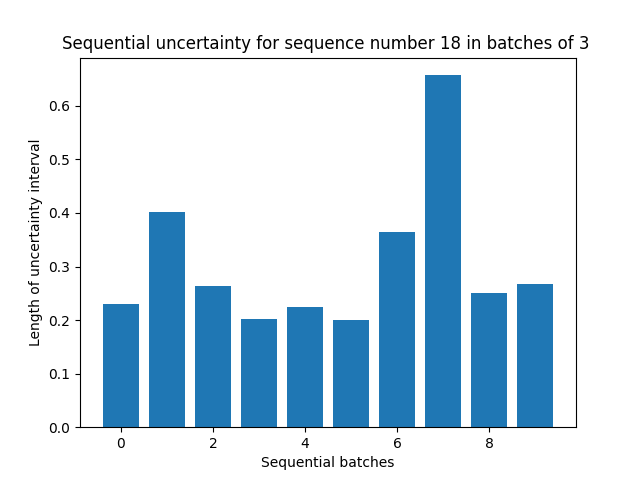} \label{fig:corr2_uncertainty}} \quad
  \subfloat[MAE]{\includegraphics[scale=0.24]{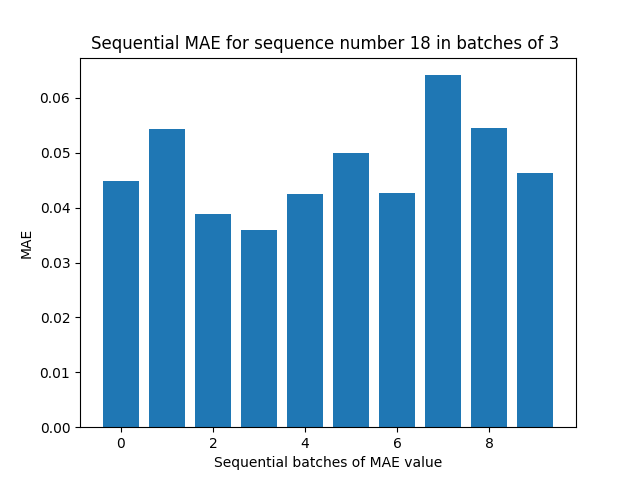}  \label{fig:corr2_mae}} 
 
  \end{figure}

Also, Figure \ref{fig:mae_corr2} displays averaged AD and PIC for a batch of three for a sequence of events corresponding to a user of \textit{Music} dataset. Here also, it is evident that the correlation between AD and prediction interval length.

\paragraph{Sensitivity Analysis}
We also perform sensitivity analysis on crime dataset for different dropouts. The results are reported in Table~\ref{Tab:sensitivitydropout}. We observe that smaller values of dropout probabilities will not change the results much. Though we can obtain a very high PIC value of 0.999 with a drop probability of $0.8$, this is  not  the best setting due to very high MAE. So, the algorithm is producing wider intervals to cover many events, however the MAE is increasing. On the contrary, dropout 0.1 is producing low MAE, however PIC is also low. So, actual event coverage will be low for such a model. 

\begin{figure}
    \centering
    \includegraphics[scale=0.20]{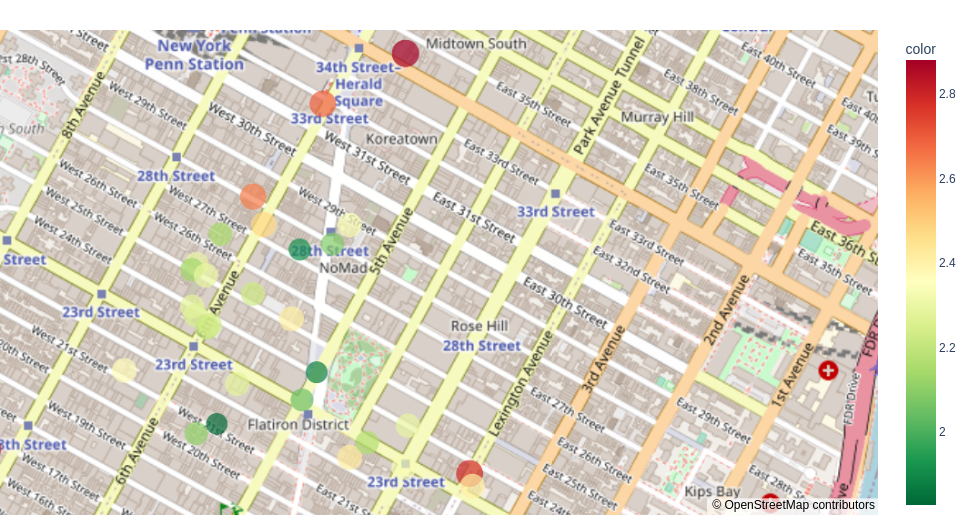}
    \caption{Plots for locations from Foursquare with color coding showing distance between predicted lower and upper bound(in km)}
    \label{fig:spatial_taxi_density}
    \vspace{-4mm}
\end{figure}

\paragraph{Spatial Analysis}
The proposed approach outputs spatial uncertainty in terms of lower and upper bound on latitude and longitude. To analyze spatial uncertainty in the units of distance, we calculate the distance between the predicted latitude-longitude bounds (PIL) for  locations associated with a user  from the Foursquare dataset and plot in Figure \ref{fig:spatial_taxi_density}. Also, Table \ref{Tab:corr_spatial} lists the correlation between predictive variance and absolute deviation (AD) for predicted time, latitude and longitude. The positive correlation coefficient indicates that  predictive uncertainty  increases with increase in AD.  

 
  
\section{Conclusion}
We propose \textit{Bayesian Neural Hawkes Process} for  predicting event occurrence times and modelling uncertainty. BNHP combines the advantages of NHP and Bayesian learning, resulting in improved predictive performance and uncertainty modelling capability. The proposed model uses a practical approach based on MC Dropout and extends it for the NHP model. We further develop BNHP for modelling spatio-temporal data.  Our experiments on simulated and real datasets demonstrate the efficacy of the proposed approach. 

\vspace{5mm}
\begin{minipage}[t]{\textwidth}
\hspace{1cm}  \begin{minipage}[b]{0.40\textwidth}

 \begin{tabular}[t]{|c|c|c|}
\hline
\textbf{Dropout} & \textbf{MAE}   & \textbf{PIC}   \\ \hline
0.01    & 0.201 & 0.294 \\ \hline
0.1     & 0.193 & 0.227 \\ \hline
0.6     & 0.203 & 0.737 \\ \hline
0.8     & 5.57  & 0.999 \\ \hline
\end{tabular}
\captionof{table}{Sensitivity Analysis for \\different dropouts for Crime dataset}
\label{Tab:sensitivitydropout}
  \end{minipage}
  \begin{minipage}[b]{0.40\textwidth}
  \centering
       \begin{tabular}[t]{|c|c|c|} 
\hline
   & \textbf{Foursq} & \textbf{Taxi}   \\ 
\hline
Time      & 0.91      & 0.75  \\ 
\hline
Lat  & 0.17      & 0.01  \\ 
\hline
Long & 0.18      & 0.01  \\
\hline
\end{tabular}
\captionof{table}{Correlation between MAE and PIL for spatial coordinates. }
\label{Tab:corr_spatial}
    \end{minipage}
  \end{minipage}

\nocite{tick}

\bibliographystyle{named}
\bibliography{ijcai22}

\end{document}


\appendix
\begin{center}
\Large
    \textbf{Supplementary  :  Bayesian Neural Hawkes Process for Event Uncertainty Prediction}
\end{center}

\section{Bayesian Neural Hawkes Process Algorithm}

Bayesian neural Hawkes process (BNHP) learns the weight matrices (variational parameters) by minimizing the following objective function (variational lower bound)
\begin{equation}
\label{eqn:supp1}
\begin{split}
& - \sum_{i=1}^N \big( \log (\frac{ \partial}{\partial \tau} \Phi(\tau_i |\boldsymbol{h}_{i-1}, t_{i}; \hat{W}_i) )
 - \Phi(\tau_i|\boldsymbol{h}_{i-1}, t_{i}; \hat{W}_i) \big) \\
 & + \lambda ( \parallel \bar{W}_t \parallel^{2} +  \parallel \bar{V}_r \parallel^{2} +   \parallel \bar{U}_r \parallel^{2}  +  \parallel \bar{b}_r \parallel^{2})
 \end{split}
\end{equation}
After learning the parameters $\bar{W}  = \{\bar{W}_t, \bar{V}_r, \bar{U}_r, \bar{b}_r\}$, predicting the future event occurrence time is done as described in  Algorithm~\ref{alg:algorithm1}, where $\hat{W}_s$ denotes a weight vector sample obtained using dropout at test time. 

\begin{algorithm}
\begin{algorithmic}[1]
\State \textbf{Input:} $\{t_1, t_2, t_3, \ldots, t_p\}$, number of MC samples $S$, parameters $W$ and neural HP network $f^W(\cdot)$ which consider previous $M$ inter-arrival times. 
\State\textbf{Output:} Time of occurrence of future predictions $\hat{t}_{p+1}, \hat{t}_{p+2}, \ldots, \hat{t}_{n} $ and uncertainty intervals as $\{\hat{t}_{p+1}^L$, $\hat{t}_{p+1}^U\}$, $\{\hat{t}_{p+2}^L$, $\hat{t}_{p+2}^U\}$ and $\{\hat{t}_{n}^L$, $\hat{t}_{n}^U\}$
\State Train the model to learn $\bar{W}$ from the train data $\{t_1, t_2, t_3, \ldots, t_p\}$ by minimizing the loss \eqref{eqn:supp1}. 
\For{$i\gets p+1,n$}
\For{$s \gets 1, S$}
\State $\Phi_i^{s}(\tau) \gets f^{(\hat{W}_s)}(\{\tau_k,t_k\}_{k=i-M}^{i-1})$
\State $\lambda_i^{s}(\tau) \gets \frac{\partial}{\partial \tau} \Phi_i^{s}(\tau)$ 
\State $\bar{t}_{i}^{(s)} \gets \textsc{Bisection}(\Phi_i^{s}(\tau))$ 
\EndFor
\State $\hat{t}_{p} \gets \frac{1}{S} \sum_{s=1}^S \bar{t}_{i}^{(s)} $ 
\State $\sigma_p^2 \gets \frac{1}{S} \sum_{s=1}^S (\bar{t}_{p}^{(s)} - \hat{t}_{p})^2$ 
\State $\hat{t}_{p}^L = \hat{t}_{p} - k*\sigma_p$,  $\hat{t}_{p}^U = \hat{t}_{p} + k*\sigma_p$
\EndFor
\end{algorithmic}
\caption{Proposed Approach}
\label{alg:algorithm1}
\end{algorithm}

\section{Dataset Details}
In this section, we describe the details of the dataset used for our setup. Dataset statistics are mentioned in Table \ref{Tab:dataset_stats}. We perform the following preprocessing for the real world datasets for our experimental setup. 
    \textbf{1)} \textbf{Crime:} This dataset contains the records of the police department calls for service in San Francisco~\footnote{\url{https://catalog.data.gov/dataset/police-calls-for-service}}. 
    Each record contains the crime and timestamp associated with the call along with other information. We have selected ten such crimes as \textit{Burgalary, Drugs, Fraud, Injury, Accident, Intoxicated Person, Mentally disturbed, Person with knife, Stolen Vehicle, Suicide Attempt, Threats/Harrassment} and their associated timestamps. Each sequence corresponds to a crime.
    \textbf{2)} \textbf{Music:} This dataset contains the history of music listening of users at \textit{lastfm}~\footnote{\url{https://www.dtic.upf.edu/~ocelma/MusicRecommendationDataset/lastfm-1K.html}}. We consider 20 sequences from 20 most active users in the month of January, 2009. Each sequence corresponds to one user.
   \textbf{3)} \textbf{New York Taxi:} The dataset contains trip records for taxis in New York City. Each pick-up record is considered as an event. This datset contains spatial attributes in the form of latitude-longitude pair along with time of occurrence of events. We consider 100,000 events for each user and a sequence is constructed for each vendor.
    \textbf{4)} \textbf{Foursquare:} This is a location-based social network which contains information about check-ins of user. We have considered top 200 users during the period April 2012 to February, 2013. Each sequence corresponds to each user. This dataset also contains spatial attributes as latitude-longitude along with time of occurrence of events. 
    \begin{table}
\centering
\caption{Dataset Statistics}
\label{Tab:dataset_stats}
\begin{tabular}{|c|c|} 
\hline
Dataset     & \# events  \\ 
\hline
Sim-Poisson & 80,000     \\ 
\hline
Sim-Hawkes  & 80,000     \\ 
\hline
Crime       & 217,662    \\ 
\hline
Music       & 112,610    \\ 
\hline
NYC         & 200,000    \\ 
\hline
Foursquare  & 95,702     \\
\hline
\end{tabular}
\end{table}

\section{Implementation Details}
Herein, we present training details for all our experiments.  We split our dataset into training, validation and test set as 70-10-20. The split is made across all sequences. Therefore, the first 70\%, next 10\% and last 20\% of the events in each sequence are used for training, validation and testing respectively. The best hyperparameters are selected using the validation set. We consider a  recurrent neural network with one layer and 64 units and  5-layer feed-forward neural network with 16 units in each layer. Recurrent Neural Network is associated with a dropout at input level as well as recurrent layer level. Along with that, there is a dropout associated with feedforward neural network to represent cumulative hazard function. Also, we use \textit{Tanh} activation function. We use Adam optimizer with learning rate, $\beta_1$ and $\beta_2$ as 0.0001, 0.90 and 0.99 along with L2 regularization. We have used 50 different architectures using Monte Carlo Dropout. The results are reported after averaging the results after running the model three times. More architecture details for hyperparameter settings is mentioned in the the supplementary. We perform 1-step lookahead prediction where we use actual time of occurrence of events as past events for the historical information to predict future event. We have perfomed all the experiments on Intel(R) Xeon(R) Silver 4208 CPU @ 2.10GHz, GeForce RTX 2080 Ti GPU and 128 GB RAM. We have implemented our code in Tensorflow 1.5.0 \cite{tensorflow2015-whitepaper}.

      \begin{figure}[t]
  \centering
  \caption{Correlation between MAE and PIL}
 \label{fig:mae_corr2}
  \subfloat[PIL] {\includegraphics[scale=0.24]{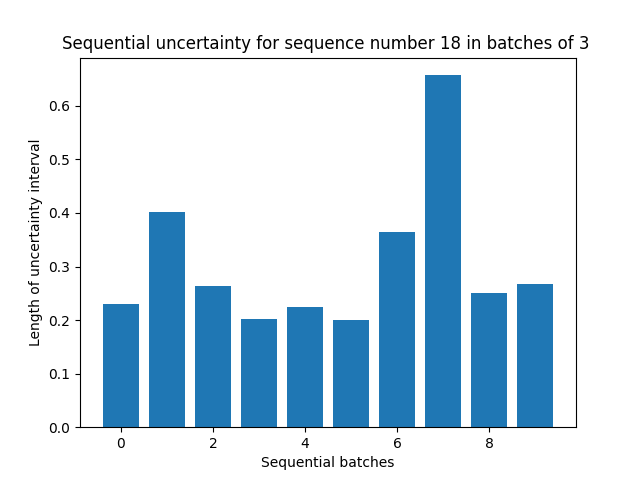} \label{fig:corr2_uncertainty}} \quad
  \subfloat[MAE]{\includegraphics[scale=0.24]{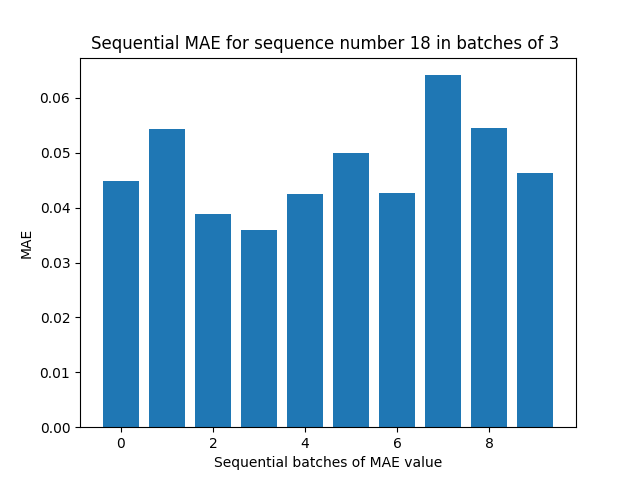}  \label{fig:corr2_mae}} 
 
  \end{figure}

\section{Correlation Analysis}
Also, Figure \ref{fig:mae_corr2} displays averaged AD and PIC for a batch of three for a sequence of events corresponding to a user of \textit{Music} dataset. Here also, it is evident that the correlation between AD and prediction interval length.

   \begin{table*}
\centering
\caption{Hyperparameter settings for the experimental set-up}
\label{Table:hyperparameter}
\begin{tabular}{|c|c|} 
\hline
\begin{tabular}[c]{@{}c@{}}Hyperparameters\\ \\ trials\end{tabular}              & \begin{tabular}[c]{@{}c@{}}Layers for representing cumulative hazard function = [2,5]\\ Dropout = [0.01,0.5,0.8,0.9]\\ Recurrent dropout = [0.01,0.1,0.5]\\ RNN Input dropout = [0.01,0.1,0.5]\\ Regularization constant = [0.001]\\ Batch size = [128,256,512,1024]\\ Truncation depth = [10,20,40,100]\end{tabular}                                                                                                                                                                                                       \\ 
\hline
\begin{tabular}[c]{@{}c@{}}Hyperparameters\\ for reported\\ results\end{tabular} & \begin{tabular}[c]{@{}c@{}}Layers for representing cumulative hazard function = 5\\ Dropout = 0.5\\ Regularization constant = 0.001\\ Seed = 11\\ Truncation depth = 20\\ Recurrent dropout = \{im-Poisson: 0.1, Sim:Hawkes: 0.1, Crime: 0.1, Music: 0.1\}\\ RNN Input dropout = \\ \{im-Poisson: 0.1, Sim:Hawkes: 0.1, Crime: 0.1, Music: 0.1\}\\ Batch size = \\ \{im-Poisson: 512, Sim:Hawkes: 512, Crime: 512, Music: 1024\}\\ Epochs = \\ \{im-Poisson: 500, Sim:Hawkes: 1500, Crime: 500, Music: 1500\}\end{tabular}  \\
\hline
\end{tabular}
\end{table*}

\bibliographystyle{named}
\bibliography{ijcai22}